\documentclass[10pt,twocolumn,letterpaper]{article}

\usepackage[pagenumbers]{cvpr} % To force page numbers, e.g. for an arXiv version

\usepackage[utf8]{inputenc} % allow utf-8 input
\usepackage[T1]{fontenc}    % use 8-bit T1 fonts
\usepackage{url}            % simple URL typesetting
\usepackage{booktabs}       % professional-quality tables
\usepackage{amsfonts}       % blackboard math symbols
\usepackage{nicefrac}       % compact symbols for 1/2, etc.
\usepackage{microtype}      % microtypography
\usepackage{xcolor}         % colors

\usepackage{amsmath}
\usepackage{multirow}
\usepackage{tabulary}
\newcolumntype{x}[1]{>{\centering\arraybackslash}p{#1pt}}

\usepackage{tikzducks}
\usepackage{graphicx}
\usepackage{xspace}
\usepackage{amssymb}% http://ctan.org/pkg/amssymb
\usepackage{pifont}% http://ctan.org/pkg/pifont
\usepackage{subcaption}
\usepackage[safe]{tipa}

\usepackage{footmisc}
\usepackage{paralist}
\usepackage[shortlabels]{enumitem}
\setlist[enumerate]{nosep}

\newcommand*\samethanks[1][\value{footnote}]{\footnotemark[#1]}
\definecolor{cvprblue}{rgb}{0.21,0.49,0.74}
\usepackage[pagebackref,breaklinks,colorlinks,allcolors=cvprblue]{hyperref}

\def\paperID{xxxx} % *** Enter the Paper ID here
\def\confName{xxxx}
\def\confYear{xxxx}

\newcolumntype{L}[1]{>{\raggedright\let\newline\\\arraybackslash\hspace{0pt}}m{#1}}
\newcolumntype{C}[1]{>{\centering\let\newline\\\arraybackslash\hspace{0pt}}m{#1}}
\newcolumntype{R}[1]{>{\raggedleft\let\newline\\\arraybackslash\hspace{0pt}}m{#1}}

\newcommand{\app}{\raise.17ex\hbox{$\scriptstyle\sim$}}

\newlength\savewidth\newcommand\shline{\noalign{\global\savewidth\arrayrulewidth
  \global\arrayrulewidth 1pt}\hline\noalign{\global\arrayrulewidth\savewidth}}
\newcommand{\tablestyle}[2]{\setlength{\tabcolsep}{#1}\renewcommand{\arraystretch}{#2}\centering\footnotesize}

\newcommand{\eqn}[1]{Eq.~(\ref{#1})}
\newcommand{\fig}[1]{Fig.~\ref{#1}}
\newcommand{\sect}[1]{Sec.~\ref{#1}}
\newcommand{\tbl}[1]{Tab.~\ref{#1}}

\definecolor{darkgreen}{RGB}{114,210,115}
\definecolor{lime}{RGB}{34,139,34}
\definecolor{gucolor}{RGB}{50,200,0}

\makeatletter
\DeclareRobustCommand\onedot{\futurelet\@let@token\@onedot}
\def\@onedot{\ifx\@let@token.\else.\null\fi\xspace}
\def\eg{\emph{e.g}\onedot} 
\def\ie{\emph{i.e}\onedot}

\def\vs{\emph{vs}\onedot}
\def\wrt{w.r.t\onedot}

\def\etal{\emph{et al}\onedot}
\makeatother

\newcommand{\cmark}{\ding{51}}%
\newcommand{\xmark}{\ding{55}}%

\newcommand{\refp}[0]{\mathbf{P}_{cam}}
\newcommand{\qryp}[0]{\mathbf{Q}_{cam}}

\newcommand{\refpg}[0]{\mathbf{P}_{G}}
\newcommand{\qrypg}[0]{\mathbf{Q}_{G}}
\newcommand{\refpgc}[0]{\mathbf{P}_{G}^c}
\newcommand{\qrypgc}[0]{\mathbf{Q}_{G}^c}
\newcommand{\refpf}[0]{\mathbf{P}^f}

\newcommand{\refpl}[0]{\mathbf{P}_{L}^f}
\newcommand{\qrypl}[0]{\mathbf{Q}_{L}^f}

\newcommand{\diffp}[0]{\Delta \mathbf{T}}
\newcommand{\diffr}[0]{\Delta \mathbf{R}}
\newcommand{\difft}[0]{\Delta \mathbf{t}}
\newcommand{\arbop}[0]{\text{AR}_{\text{BOP}}}

\newcommand{\netname}[0]{UNOPose}

\newcommand{\rrot}[0]{\mathbf{r}}
\newcommand{\rot}[0]{\mathbf{R}}
\newcommand{\trans}[0]{\mathbf{t}}
\newcommand{\size}[0]{s}

\newcommand{\nbf}[1]{{\noindent \textbf{#1}}}

\title{UNOPose: Unseen Object Pose Estimation with \\ an Unposed RGB-D Reference Image}

\author{Xingyu Liu$^{1,}$\,\thanks{Equal contributions.}~,
~~~Gu Wang$^{1,}$\,\samethanks~,
~~~Ruida Zhang$^{1}$, \\
 Chenyangguang Zhang$^1$,
~~~Federico Tombari$^{2,\,3}$,
~~and Xiangyang Ji$^{1}$ \\
\textsuperscript{1}Tsinghua University,~~\textsuperscript{2}Technical University of Munich,
~~\textsuperscript{3}Google \\
{\tt\small \{liuxy21@mails., wanggu1@, zhangrd23@mails., zcyg22@mails., xyji@\}tsinghua.edu.cn,} \\
{\tt\small tombari@in.tum.de}
}

\begin{document}
\maketitle
% abstract =====================================
\begin{abstract}
Unseen object pose estimation methods often rely on CAD models or multiple reference views, making the onboarding stage costly.
To simplify reference acquisition,
we aim to estimate the unseen object's pose through a single unposed RGB-D reference image.
While previous works leverage reference images as pose anchors to limit the range of relative pose,
our scenario presents significant challenges since the relative transformation could vary across the entire $SE(3)$ space. 
Moreover, factors like occlusion, sensor noise, and extreme geometry could result in low viewpoint overlap.
To address these challenges, we present a novel approach and benchmark, termed \netname\footnote{UNO (\textipa{/"u:noU/}) means one in Spanish and Italian.}, for \underline{UN}seen \underline{O}ne-reference-based object \underline{Pose} estimation.
Building upon a coarse-to-fine paradigm, \netname~constructs an $SE(3)$-invariant reference frame to standardize object representation despite pose and size variations.
To alleviate small overlap across viewpoints, we recalibrate the weight of each correspondence based on its predicted likelihood of being within the overlapping region.
Evaluated on our proposed benchmark based on the BOP Challenge, \netname~demonstrates superior performance, significantly outperforming traditional and learning-based methods in the one-reference setting and remaining competitive with CAD-model-based methods.
The code and dataset are available at \href{https://github.com/shanice-l/UNOPose}{github.com/shanice-l/UNOPose}.

\end{abstract}
\section{Introduction}

Localizing an object in Euclidean space by estimating its 6DoF pose, \ie, 3DoF orientation and 3DoF position, plays a crucial role in augmented/virtual reality~\cite{marchand2015pose,su2019arvr}, scene understanding~\cite{nie2020total3dunderstanding,huang2018cooperative} and robotic manipulation~\cite{tremblay2018deep,du2021vision,fu2024lanpose}.
The vast majority of works~\cite{li2019cdpn,labbe2020cosypose,peng2019pvnet,wang2019densefusion,Wang_2021_GDRN,Wang_2021_self6dpp,tang2024fafa,wang2020self6d} focuses on instance-level object pose estimation, where the training and testing datasets consist of an identical set of known object instances, and the CAD models of objects are often required for generating training images and labels \cite{Denninger2023_blenderproc2,liu2024rasim}.
More recently, Wang \etal \cite{wang2019nocs} and its successors~\cite{Tian_ECCV20_DeformNet,Chen_CVPR20_CASS,Wang_IROS21_Cascaded} extended this paradigm to the category level, intending to estimate poses for novel instances within predefined %canonical-pose-aligned 
categories without requiring CAD models of target objects.
Both paradigms have limitations in open-world applications,
where annotating and training for new objects beyond known categories would be very labor-intensive and sometimes prohibitive.
% where annotating and training for novel objects or categories would be very labor-intensive.

\begin{figure*}[t]
\centering
\includegraphics[width=1.0\textwidth]{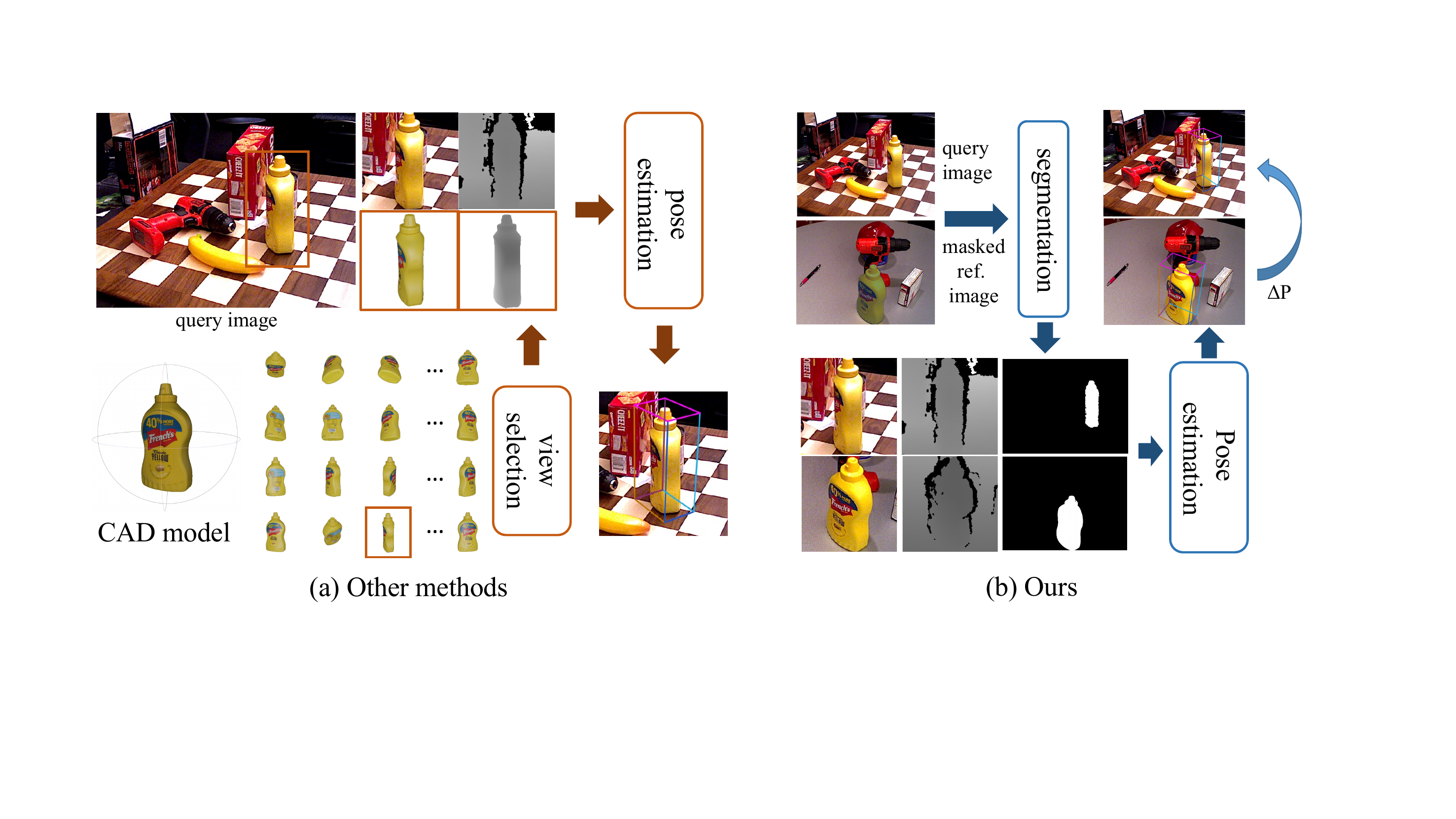}
\caption{\textbf{Illustration of unseen object pose estimation.} Given a query image presenting a target object unseen during training, we aim to estimate its segmentation and 6DoF pose \wrt a reference frame. While previous methods~\cite{labbe2023megapose,foundationposewen2024,liu2022gen6d,sun2022onepose} often rely on the CAD model or multiple RGB(-D) images for reference, we merely use one unposed RGB-D reference image.}
\label{fig:teaser}
\vspace{-5mm}
\end{figure*}

To mitigate this problem, recent research efforts \cite{foundationposewen2024,labbe2023megapose,lin2023sam6d,cai2022ove6d} have shifted towards pose estimation of arbitrary novel objects, where the target object is unseen during training. 
This task presents a great challenge due to the inherent uncertainty of the novel object's canonical frame (defined in the CAD model of the object).
Current approaches~\cite{liu2022gen6d,labbe2023megapose,nguyen2024gigaPose} often tackle this by using omnidirectional reference views to cover the target object, separating pose estimation task to viewpoint selection and relative pose estimation to known pose anchors.
If CAD models are available during the test phase, reference views can be easily produced through image rendering \cite{foundationposewen2024,labbe2023megapose,lin2023sam6d,nguyen2023cnos}.
Alternatively, without CAD models, reference views can be obtained by capturing multiple images of the object \cite{sun2022onepose,he2022onepose++,he2022fs6d,liu2024gfreedet}.
Both setups have shortcomings, as creating a CAD model of an object or labeling a substantial number of object poses is limited in scalability.
For example, they can hardly adapt to in-the-wild scenarios where novel objects are introduced unpredictably.
% can be exceedingly time-consuming.

To sum up, while existing methods exhibit promising transfer capability to novel objects, they are notably constrained by their reliance on the CAD model or multiple reference views (Fig.~\ref{fig:teaser} (a, b)).
% In this work, we address the challenge of estimating novel object poses using a single unposed RGB-D reference image (Fig.~\ref{fig:teaser}).
In this work, we aim to minimize 
% preparatory efforts
onboarding cost by estimating novel object poses using a single unposed RGB-D reference image (Fig.~\ref{fig:teaser}).
This setup is different from existing methods mainly in two aspects.
First, we focus on estimating the relative pose between two unposed viewpoints rather than the absolute pose.
Note that the absolute pose of a novel object becomes ill-posed without a certain canonical frame, while relative pose estimation remains well-defined regardless of the canonical frame's definition.
Secondly, some existing relative pose estimation methods~\cite{zhao2024dvmnet,zhao20233d,zhang2022relpose} merely use RGB modality to predict the 3DoF rotation, limiting their ability to predict relative translation and generalize beyond their training datasets.
In contrast, we fully leverage RGB-D modality to predict the 6DoF relative pose, meanwhile enhancing the network's generalizability to novel objects and environments.

Nevertheless, estimating the pose of unseen objects with only one reference view presents significant challenges from various aspects.
First, previous methods~\cite{cai2022ove6d,labbe2023megapose} select the most similar reference view as a pose anchor, distinctly reducing the search space for relative poses.
However, in our scenario, where both the target and reference object poses are unknown, the relative pose can vary across the entire $SE(3)$ space.
Moreover, in partial-to-partial object matching, factors such as occlusion, sensor noise, and extreme geometry can severely interfere with the matching process.
To address these challenges, we propose a novel approach and benchmark for \emph{UN}seen \emph{O}ne-reference-based object \emph{Pose} estimation (\emph{\netname}).
Our pipeline leverages the strong generalization capabilities of vision foundation models~\cite{oquab2023dinov2,kirillov2023segment} to produce an effective segmentation of the unseen object. 
Then, \netname~employs a coarse-to-fine % correspondence matching 
paradigm for estimating the relative pose between the reference and the query objects.
To alleviate the challenge of diverse pose and size 
variations, we introduce the $SE(3)$-invariant global reference frame (\textit{GRF}) to standardize object representation.
Subsequently, a hierarchical object encoding paradigm based on the local reference frame (\textit{LRF}) further captures fine-grained geometric details.
For achieving reliable correspondences in partial-to-partial object matching,
we harness an overlap predictor to identify and concentrate on the overlapping region.

Moreover, we propose a new benchmark devised from the BOP challenge \cite{hodan2024bop} to facilitate evaluation and future research of unseen object pose estimation with one reference.
Extensive experiments on YCB-V~\cite{posecnn}, LM-O~\cite{brachmann2014learning} and TUD-L~\cite{hodan2018bop} datasets demonstrate that our \netname~surpasses all compared methods on a single reference setting.
To our surprise, \netname~with an unposed reference is even on par with some \(SE(3)\)-invariant-feature-based methods relying on CAD models (Ours 70.9\% \vs ZTE-PPF~\cite{zteppf} 69.0\% \vs Koenig-PPF~\cite{konig2020hybrid} 75.1\% \wrt $\arbop$ metric).

Our contributions can be summarized as follows:
\begin{compactitem}
\item %
% 1) 
To the best of our knowledge, we are the first to conduct unseen object 6DoF pose estimation leveraging a single unposed RGB-D reference.
\item %
% 2) 
Based on the BOP Challenge, we devise a new extensive benchmark tailored for unseen object segmentation and pose estimation with one reference.
Additionally, we evaluate several traditional and learning-based methods on this benchmark for completeness. 
\item %
% 3) 
We introduce \netname, a network for learning relative transformation between reference and query objects. 
To achieve this, we propose the $SE(3)$-invariant global and local reference frames, enabling standardized object representations despite variations in pose and size.
Furthermore, the network can automatically adjust the confidence of each correspondence by incorporating an overlap predictor.
\end{compactitem}
\section{Related Work}

\nbf{Class-specific Pose Estimation.}~~~
Class-specific object pose estimation aims at predicting 6DoF poses of either instance of a known object (instance-level) or unseen objects within a known category (category-level).
% with the aid of CAD models.
Instance-level object poses are often solved by direct regression~\cite{Wang_2021_GDRN,li20deepim_ijcv,hu2020single,liu2024rasim} using deep neural networks,
or by establishing 2D-3D~\cite{li2019cdpn,peng2019pvnet,haugaard2022surfemb} or 3D-3D correspondences~\cite{shugurov2021dpodv2,he2021ffb6d,wang2019densefusion} which are then leveraged by RANSAC-based PnP/Kabsch algorithms.
% To address the limitations of the instance-level setting,
The instance-level setting requires expensive data generation/annotation \cite{uy2021joint,zhang2024kpred,di2023ured,di2024shapematcher}
and training for every new object.
To address this limitation, category-level pose estimation~\cite{wang2019nocs} is proposed to estimate 9DoF poses of novel objects among specific categories without CAD models. % As an aid, most methods~\cite{Tian_ECCV20_DeformNet,chen_ICCV21_sgpa,ze2022wild6d} use an intra-class mean shape as prior knowledge.
Mainstream approaches can also be categorized into direct regression~\cite{Chen_CVPR20_CASS,Lin_ICCV21_DualPoseNet,Chen_CVPR21_FSNet,di2022gpv,zhang2022rbp,liu_2022_catre,zhang2022ssp,zhang2024lapose,chen2024secondpose} or correspondence-based~\cite{wang2019nocs,chen_ICCV21_sgpa,Tian_ECCV20_DeformNet,fan2022object} methods.
However, this setting remains constrained to % a limited set of
a limited number of categories, given the additional challenge of aligning canonical frames, %for each category
managing symmetric objects, % and the ambiguity arising from similar objects %belonging to distinct categories.
% categorized differently.
and categorizing objects in similar categories correctly (\eg can and bottle).

\nbf{Novel Object Pose Estimation.}~~~
Estimating the pose of novel objects beyond known classes, \ie, objects at inference time are unseen during training, is a useful yet
challenging task.
Existing works solve this problem through image-based matching~\cite{cai2022ove6d,labbe2023megapose,foundationposewen2024,shugurov2022osop,nguyen2022templates,zhao2022fusing,nguyen2023cnos} or feature-based matching approaches~\cite{sun2022onepose,he2022onepose++,ornek2023foundpose,andera2023freeze,nguyen2024gigaPose,huang2024matchu,lin2023sam6d}.
Notably, given a target image, OVE6D~\cite{cai2022ove6d} and MegaPose~\cite{labbe2023megapose} retrieve the most similar viewpoint from a pre-rendered omnidirectional image database as a coarse pose estimate.
A customized neural network then refines the coarse pose estimate.
FoundationPose~\cite{foundationposewen2024} further builds a pose ranking network to score each refined pose hypothesis.
Feature-matching methods learn local feature descriptors to construct pixel-level or point-level correspondences.
For example, OnePose~\cite{sun2022onepose} and OnePose++~\cite{he2022onepose++} reconstruct the 3D point cloud of an unseen object to establish 2D-3D correspondences, 
whereas SAM-6D \cite{lin2023sam6d} builds 3D-3D correspondences. 
More recently, some methods~\cite{ornek2023foundpose,andera2023freeze} directly exploit the power of foundation models for zero-shot pose estimation.
Exemplarily, FoundPose~\cite{ornek2023foundpose}
extracts DINOv2 visual features~\cite{oquab2023dinov2}, while FreeZe~\cite{andera2023freeze}
further utilizes generalizable geometric features~\cite{poiesi2022gedi} to conduct feature matching.

\nbf{Relative Pose Estimation.}~~~
Orthogonal to previous works which rely on known instances or multiple reference views, relative object pose estimation estimates a relative transformation of the unseen object with only one reference image.
One closely related topic is camera pose estimation from two views.
For example, RelPose~\cite{zhang2022relpose} and RelPose++~\cite{lin2023relpose++} infer a distribution of relative rotations by leveraging an energy-based formulation.
Moreover, iFusion~\cite{wu2023ifusion} optimizes the unknown relative pose by inverting the novel view synthesis diffusion model~\cite{liu2023zero123}.
Relative camera pose estimation relies on static background to establish correspondence between viewpoints.
In the field of relative object pose estimation,
%POPE
given two RGB images, 3DAHV~\cite{zhao20233d} presents a hypothesis-and-verification framework to score each relative pose hypothesis.
Extended by that, DVMNet~\cite{zhao2024dvmnet} introduces a hypothesis-free pipeline to compute relative poses via deep voxel matching.
%
% The works mentioned above can only infer a 3DoF relative rotation transformation.
The above works have limited capability in inferring 3DoF relative translations.
In this work, with the introduction of depth data, we can estimate a complete 6DoF pose including rotation and translation.

\nbf{Point Cloud Registration.}~~~
% Our work follows the paradigm of correspondence-based methods~\cite{deng2018ppffoldnet,deng2018ppfnet,wang2019deep,li2020idam}.
% These methods first extract 3D local feature descriptors to build correspondences between source and target point clouds, and then estimate the relative pose with SVD or RANSAC.
Our approach follows the paradigm of correspondence-based methods~\cite{deng2018ppffoldnet,deng2018ppfnet,wang2019deep,li2020idam}, which initially extract 3D local feature descriptors to establish correspondences between source and target point clouds.
Subsequently, they estimate the relative transformation using techniques like SVD, RANSAC, or Hough Voting~\cite{DHVR_LKCP21_ICCV}.
% Some rotational invariant feature descriptors are proposed to eliminate the influence of canonical pose.
Some feature descriptors are proposed to ensure rotational invariance.
For example, FPFH~\cite{rusu2009fast} descriptor is computed with geometric surface properties.
PPF~\cite{drost2010model} uses the distance and angles to describe the relation of point pairs.
TOLDI~\cite{yang2017toldi} proposes a robust local reference frame using keypoint normals and neighboring projections.
With the advent of deep learning, neural networks are increasingly employed to extract 3D local or global feature descriptors~\cite{wang2019deep,huang2021predator,qin2023geotransformer,choy2019fcgf,RIGA_YHQS*24_tpami}.
%Point cloud registration has broad applications in object pose estimation with CAD models~\cite{lee2021deeppro,dang2022learning,zhao2023learning,lin2023sam6d}.

\begin{figure*}[t]
\centering
\includegraphics[width=1.0\textwidth]{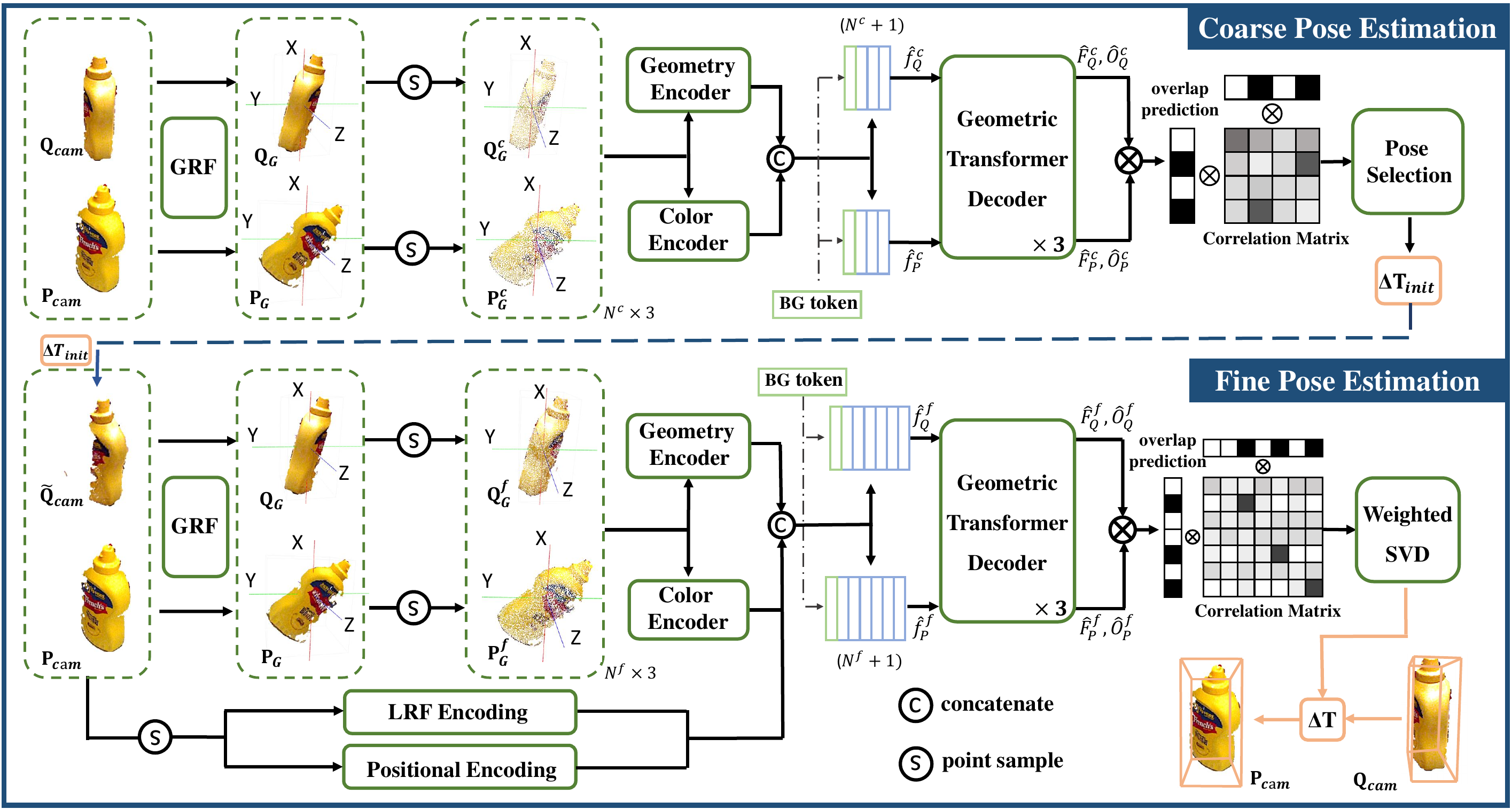}
\caption{\textbf{The network architecture of \netname.}
%We leverage a coarse-to-fine paradigm and hierarchically build correlation matrices. 
Given the query and reference point clouds $\qryp$ and $\refp$ in the camera frame, \netname~first transforms them into the $SE(3)$-invariant global reference frame (GRF). 
% Geometry and color encoders then extract feature descriptors from coarse point sets, which are subsequently used to build coarse correlation matrices.
Then feature descriptors are extracted from sparse point sets for constructing the coarse correlation matrix.
For achieving precise correspondences, the fine pose estimation module exploits structural details using positional encoding and local reference frame (LRF) encoding.
}
\label{fig:network}
\vspace{-3mm}
\end{figure*}

\section{UNO Object Segmentation and Pose Estimation}
\label{sec:UNO_method}
% This section presents the pipeline of solving unseen object poses from one reference.
% We first address the problem formulation in \sect{sec_probfrom}, and then introduce one-reference-based unseen object segmentation (\sect{sec_objseg}) and pose estimation (\sect{sec_objpose}) consecutively.

\subsection{Problem Formulation}
\label{sec_probfrom}
% Given a masked RGB-D reference image containing target object $\mathcal{O}$, our goal is to estimate a 6DoF relative transformation [$\rot_\delta$ | $\trans_\delta$] between reference and target object.
Assuming an arbitrary unseen rigid object in a query image, our goal is to estimate the object's mask $M_q$ and its 6DoF relative pose $\diffp \in SE(3)$ with a masked RGB-D reference image exhibiting the target object without major occlusion or truncation.
As illustrated in \fig{fig:teaser}, the input is: %\\ % as follows: \\
\begin{compactenum}[1)]
% 1) $I_q \in \mathbb{R}^{H \times W \times 3}$ and $D_q \in \mathbb{R}^{H\times W}$: The query RGB image and depth image; \\
\item $[I_q|D_q] \in \mathbb{R}^{H \times W \times 4}$: The query RGB-D image; %\\
% 2) $I_p \in \mathbb{R}^{H \times W \times 3}$ and $D_p \in \mathbb{R}^{H\times W}$: The reference RGB image and depth image; \\
\item $[I_p|D_p] \in \mathbb{R}^{H \times W \times 4}$ and $M_p \in \mathbb{R}^{H \times W}$: The reference RGB-D image and the corresponding binary mask indicating the target object; %\\
\item $K_q \in \mathbb{R}^{3 \times 3}$ and $K_p \in \mathbb{R}^{3 \times 3}$: Camera intrinsics of the query and reference images. %\\
% 4) $M_p \in \mathbb{R}^{H \times W}$: A binary mask on the reference image indicating the target object. \\
\end{compactenum}
Optionally, if the pose $\textbf{T}_p \in SE(3)$ of the reference object in the camera frame is known
% \footnote{Note that the method should not rely on the absolute pose of the reference object, since the world frame can vary arbitrarily in different applications. We only use the reference object pose for the convenience of evaluation on standard object pose datasets.}
, 
the query object pose $\mathbf{T}_q \in SE(3)$ can be recovered by $\mathbf{T}_q = \diffp \mathbf{T}_p$.
% \begin{equation}
%    \mathbf{T}_q = \diffp  \mathbf{T}_p.
% \end{equation}
Note that, for practicality, the method should not rely on the absolute pose of the reference object, since the world frame can vary arbitrarily in different applications. We only use the reference object pose % for the convenience of evaluation on standard object pose datasets. 
during the inference stage for evaluation on standard object pose datasets. 

\subsection{UNO Object Segmentation}
\label{sec_objseg}
Firstly, we segment the query object from a cluttered background.
Thanks to the great generalization ability of vision foundation models, %~\cite{kirillov2023segment,oquab2023dinov2}, 
methods like~\cite{nguyen2023cnos,lin2023sam6d,chen2024zeropose} can effectively segment novel objects with CAD models. 
However, unlike previous approaches, which generate % omnidirectional
diverse descriptors from multiple rendered views, we can access \emph{only a single reference image}. 
% To address this challenge, we fully exploit global and local descriptors from DINOv2 \cite{oquab2023dinov2} to score the SAM \cite{kirillov2023segment} mask proposals.
To address this challenge, we use the SAM model \cite{kirillov2023segment} to predict all possible mask proposals from the query image, and then score each mask proposal by comparing DINOv2 \cite{oquab2023dinov2} descriptors between the query and reference images using cosine similarity, identifying the most similar mask $M_q$.
Please refer to the appendix for more details about UNO segmentation.

\subsection{UNO Object Pose Estimation}
\label{sec_objpose}

\subsubsection{Overview of \netname}

Given the predicted mask $M_q$ of the query image and mask $M_p$ of the reference image, we crop and back-project the object of interest from depth maps $D_q, D_p$ into the camera space as two point sets $\qryp \in \mathbb{R}^{N^Q \times 3}$ and $\refp \in \mathbb{R}^{N^P \times 3}$, where $N^Q$ and $N^P$ denotes the point numbers of query and reference point clouds, respectively.
Our goal is to recover the relative transformation $\diffp = \{\diffr, \difft \}$ by 
minimizing the correspondence distance
\begin{equation}
\label{eq:target}
    \min \sum_{(\mathbf{q},~\mathbf{p}) \in \mathbf{C}} \|\diffr \mathbf{q} + \difft - \mathbf{p}\|_2,
\end{equation}
where $\mathbf{C}$ is the predicted correspondence set between $\qryp$ and $\refp$.
We exploit both color and geometric cues to build this correspondence and follow the broadly used coarse-to-fine paradigm in point cloud registration to solve \eqn{eq:target}. %hierarchically. 
The network is demonstrated in \fig{fig:network}.
\subsubsection{Coarse-to-fine Pose Estimation}
\label{sec:method_pose}

% \paragraph{Transformation with Global Reference Frame}
\paragraph{Constructing a Pose-invariant Reference Frame.}
Given only an unposed reference image, the relative pose to predict is arbitrary in the \(SE(3)\) space, which renders a significant challenge for achieving robust correspondences.
% In relative object pose estimation, the object of interest can be anywhere in space with arbitrary pose and size, which renders a significant challenge for achieving robust correspondences.
Hence, we introduce a pose-invariant global reference frame (\textbf{GRF}) and transform $\qryp$ and $\refp$ into GRF as $\qrypg$ and $\refpg$.

Concretely, taking $\qryp$ as an example,
transforming the point cloud to GRF involves a 7DoF coordinate transformation $\{\rot_G\in SO(3),\,\trans_G\in \mathbb{R}^3,\,\size_G \in \mathbb{R}\}$:
\begin{equation}
\label{grf_all}
    % \qrypg &=  \rot_G^\top (\qryp - \trans_G) / \size_G, \\
    \qrypg =  \{\rot_G^\top (\mathbf{q} - \trans_G) / \size_G \,|\, \mathbf{q}\in \qryp\}.
\end{equation}
The origin of GRF is located at the object center $\mathbf{c}_Q$ for translation invariance, and the radius of the point cloud is rescaled to 1 for size invariance, computed as
\begin{equation}
\label{grf_ts}
    \trans_G = \mathbf{c}_Q, 
    \quad 
    \size_G = \max_{\mathbf{q} \in \qryp} \| \mathbf{q} - \mathbf{c}_Q \|_2.
\end{equation}
The key ingredient is to devise the rotation $\rot_G = [\rrot_{Gx}| \rrot_{Gy}| \rrot_{Gz}]$, where $\rrot_{Gx}, \rrot_{Gy}, \rrot_{Gz}$ are the three columns of $\rot_G$, %rendering the transformed point cloud orientation-invariant.
ensuring orientation invariance for the transformed point cloud.
% , which should be orientation-invariant and only related to the spatial distribution of points.
Inspired by former works~\cite{yang2017toldi,stein1992structural,frome2004recognizing}, we use the normal vector of the object center $\mathbf{n}(\mathbf{c}_Q)$ as $\rrot_{Gz}$, and project all points in $\qryp$ onto the tangent plane of $\mathbf{c}_Q$, summarizing the projected vectors to determine $\rrot_{Gx}$.
Then, $\rrot_{Gy}$ is obtained by the cross-product of $\rrot_{Gx}$ and $\rrot_{Gz}$, \ie, $\rrot_{Gy} = \rrot_{Gx} \times \rrot_{Gz}$. 
This ensures that the XY-plane evenly divides the point cloud, with the x-axis representing the principal projection direction.
% Taking $\qryp$ as an example, for each point $\mathbf{q} \in \qryp$ and any transformation \textbf{T} in $SE(3)$:
% \begin{equation}
%     \forall \mathbf{q} \in \qryp, \mathbf{T} \in SE(3), \text{GRF}[\mathbf{T}(\mathbf{q})] = \text{GRF}[\mathbf{q}].
% \end{equation}
% This transformation renders that, regardless of the object's position, orientation, or size in the camera frame, the point clouds are represented in a way that is invariant by these factors. 
% This $SE(3)$-invariant frame standardizes object representation despite variations in pose and size.
% transformation ensures that the input to the network is consistent with the object pose, 
%, focusing solely on the spatial distribution of the points.
The GRF is $SE(3)$-invariant as it can be uniquely determined by the spatial distribution of the point cloud, making the transformed point cloud robust to variations in pose and size.

Concretely, we perform SVD on the covariance matrix $\text{Cov}(\qryp) = 
\frac{1}{N_Q}\qryp^\top\qryp - \mathbf{c}_Q \mathbf{c}_Q^{\top}$ and use the singular vector \wrt the minimum singular value
%of $\text{Cov}(\mathbf{Q})$ 
to determine $\rrot_{Gz}$:
\begin{equation}
    \rrot_{Gz} = \begin{cases}
        \mathbf{n}(\mathbf{c}_Q),~~~~~\text{if}~~\mathbf{n}(\mathbf{c}_Q)^\top \sum_{\mathbf{q} \in \qryp} (\mathbf{c}_Q - \mathbf{q}) > 0 \\
        -\mathbf{n}(\mathbf{c}_Q),~~\text{otherwise}.
     \end{cases}
     \label{gz}
\end{equation}
Afterwards, $\rrot_{Gx}$ is computed as
% Specifically, the transformation is described as a 9DoF pose $\{\rot_G,\,\trans_G,\,\size_G\}$.
% % a 9DoF pose transformation $\{\rot_G,\,\trans_G,\,\size_G\}$.
% Intuitively, we can define $\trans_G$ as the center $\mathbf{c}_Q$ and $\size_G$ as the size of the point cloud.
% The key ingredient is how to 
% %disentangle
% devise rotation $\rot_G = [\rrot_{Gx}| \rrot_{Gz} \times \rrot_{Gx}| \rrot_{Gz}]$ for object pose invariance.
% Inspired by former works~\cite{yang2017toldi,stein1992structural,frome2004recognizing}, we first perform SVD on the covariance matrix $\text{Cov}(\mathbf{Q}) = 
% \frac{1}{N_Q}\qryp^\top\qryp - \mathbf{c}_Q \mathbf{c}_Q^{\top}$ and then use the singular vector \wrt the minimum singular value of $\text{Cov}(\mathbf{Q})$, \ie, the normal of the center $\mathbf{n}(\mathbf{c}_Q)$, to determine $\rrot_{Gz}$:
% \begin{equation}
%     \rrot_{Gz} = \begin{cases}
%         \mathbf{n}(\mathbf{c}_Q),~~~~~\text{if}~~\mathbf{n}(\mathbf{c}_Q)^\top \sum_{\mathbf{q} \in \qryp} (\mathbf{c}_Q - \mathbf{q}) > 0 \\
%         -\mathbf{n}(\mathbf{c}_Q),~~\text{otherwise}.
%      \end{cases}
%      \label{gz}
% \end{equation}
% % Denoting the surface plane of $\mathbf{c}_Q$ as $E_\mathbf{c}$, we project all points in $\qryp$ to $E_\mathbf{c}$ and summarize all the projected vector as the direction of $\rot_{Gx}$, computed as
% Afterwards, we project all points in $\qryp$ to the surface plane of $\mathbf{c}_Q$ and summarize all the projected vector as the direction of $\rrot_{Gx}$, computed as
\begin{equation}
\label{rgx}
    \rrot_{Gx} = \sum_{\mathbf{q} \in \qryp} w_q\left((\mathbf{q} - \mathbf{c}_Q) - \rrot_{Gz}^\top (\mathbf{q} - \mathbf{c}_Q) \rrot_{Gz}\right),
\end{equation}
where $w_q$ is a weight \wrt distance between $\mathbf{q}$ and $\mathbf{c}_Q$ (See % Sec.~\ref{sec:details_grf} 
the appendix for details).
% Finally, with settled $\{\rot_G,\,\trans_G,\,\size_G\}$, 
% $\qrypg$ is measured by
% \begin{equation}
% \label{grf_all}
%      \qrypg =  \rot_G^\top (\qryp - \trans_G) / \size_G.
% \end{equation}
% Similarly, we transform $\refp$ to $\refpg$.

Previous works often depend on complex networks or computationally expensive PPF features for $SE(3)$ invariance~\cite{li2021leveraging,you2022cppf,yu2023rotation}.
In contrast, our transformation is computationally efficient and can seamlessly 
adapt to a wide range of network architectures.

\nbf{Coarse Pose Estimation.}~~
Given point clouds $\qrypg$ and $\refpg$ in GRF, we sample two sparse point sets $\qrypgc \in \mathbb{R}^{N^{c} \times 3}$ and $\refpgc \in \mathbb{R}^{N^{c} \times 3}$ to efficiently obtain a coarse pose initialization $\diffp_{init}$.
Specifically, we leverage a geometry encoder~\cite{qin2023geotransformer} and a color encoder~\cite{oquab2023dinov2} to extract point cloud and RGB features separately. 
Features are further concatenated as $f_{Q}^{c} \in \mathbb{R}^{N^{c} \times d} $ and $f_{P}^{c} \in \mathbb{R}^{N^{c} \times d}$, where $d$ is the dimension of embeddings.
Following~\cite{lin2023sam6d}, we add a learnable background token for assigning non-overlapping points.
The embeddings, denoted as $\hat{f}_{Q}^{c} \in \mathbb{R}^{(N^{c} + 1) \times d} $ and $\hat{f}_{P}^{c} \in \mathbb{R}^{(N^{c} +1) \times d}$, are processed by three stacked Geometric Transformer decoding modules~\cite{qin2023geotransformer}.

The outputs of the last decoder $\hat{F}_{Q}^{c} \in \mathbb{R}^{(N^{c} + 1) \times D}$ and $\hat{F}_{P}^{c} \in \mathbb{R}^{(N^{c} +1) \times D}$ are point-wise features for building the correlation matrix.
% However, in our setting, the overlap ratio between $\qrypg$ and $\refpg$ is oftentimes meager.
However, in our setting, the overlap ratio can be influenced by complicated factors like viewpoint, occlusion or depth noise.
To solve this issue, the network additionally predicts overlap confidences  $\hat{O}_{Q}^{c} \in \mathbb{R}^{(N^{c} + 1) \times 1}$ and $\hat{O}_{P}^{c} \in \mathbb{R}^{(N^{c} +1) \times 1}$, which indicate point-wise probabilities of being in the overlapping region.
The overlap-aware correlation matrix $\mathbf{X}^{c} \in \mathbb{R}^{(N^{c} + 1)\times (N^{c} + 1)}$ can thus be computed 
%from the output of the last decoder, is represented 
by
\begin{equation}
    \mathbf{X}^{c} = \texttt{softmax}[(\hat{O}_{Q}^{c} \odot \hat{F}_{Q}^{c}) 
    %\times
    (\hat{O}_{P}^{c} \odot \hat{F}_{P}^{c})^\top].
\end{equation}
Here $\odot$ indicates element-wise multiplication.
Each element in $\mathbf{X}^{c}$ suggests a correlation score between a point pair in $\mathbf{Q}^{c}$ and $\mathbf{P}^{c}$.

Once $\mathbf{X}^{c}$ is computed, we can extract all possible corresponding point pairs between $\mathbf{Q}^{c}$ and $\mathbf{P}^{c}$ as well as their correlation scores to solve \eqn{eq:target}. 
We first compute pose hypotheses by sampling $N_H$ triplets of point pairs according to the distribution of $\mathbf{X}^c$. 
Then each pose hypothesis $\diffp_h = \{\diffr_h, \difft_h\}$ is scored with the reciprocal of distance $D_h$ as in ~\cite{haugaard2022surfemb,lin2023sam6d}
\begin{equation}
\begin{split}
        D_h &= \frac{1}{N^{c}} \sum_{{\mathbf{p}^{c}} \in {\mathbf{P}^{c}}} \min_{{\mathbf{q}^{c}} \in {\mathbf{Q}^{c}}} \| {\diffr_h^\top} ({\mathbf{q}^{c}} - \difft_h ) - {\mathbf{p}^{c}}\|_2, 
        \\ 
        \xi_h &= \frac{1}{D_h},~~~h=1,2,3,...,N_H.
\end{split}
\end{equation}
%where $N_H$ is the number of pose hypotheses.
The pose hypothesis with the highest score $\xi_h$ is selected as the initial pose prediction $\diffp_{init}=\{\diffr_{init}, \difft_{init}\}$ and further sent to transform the input of the next stage.

\nbf{Fine Pose Estimation.}~~
After transforming $\qryp$ to $\mathbf{\tilde{Q}}_{cam}$ with the initial pose prediction $\diffp_{init}$,
we perform a fine matching process between two dense point sets, \ie, $\mathbf{\tilde{Q}}^{f} \in \mathbb{R}^{N^{f} \times 3}$ and $\refpf \in \mathbb{R}^{N^{f} \times 3}$, to achieve a more accurate pose.
The network exploits geometric details through a hierarchical encoding paradigm, comprising a positional encoding layer and a local reference frame encoding layer.
% Taking $\qryphat$ as an example, for each point $\mathbf{\tilde{q}}^m$ in $\qryphat$,
For each point,
we first encode its global position using a mini-PointNet~\cite{qi2017pointnet},
and then construct SE(3)-invariant local reference frame (\textbf{LRF}) to gather local descriptors.
These two encodings complement each other,
as local descriptors capture fine geometric structures within small neighborhoods, while the positional encoding layer offers global geometric context.
Leveraging $\mathbf{\tilde{Q}}^{f}$ for example, the process of constructing LRF encoding is as follows.
For each point $\mathbf{\tilde{q}}^m$ in $\mathbf{\tilde{Q}}^{f}$, 
we build a local region set $\mathcal{\tilde{Q}}^m = \{\mathbf{\tilde{q}}_{j}, \text{where}~\|\mathbf{\tilde{q}}_{j} - \mathbf{\tilde{q}}^m\|_2 \leq r\}_{j=1}^{N_D}$ by grouping $N_D$ neighboring points of $\mathbf{\tilde{q}}^m$.
% within a ball with radius $r$~\cite{qi2017pointnet++}.
The transformation pose $\{\rot_L^m,\,\trans_L^m,\,\size_L^m\}$ for LRF is computed similarly to GRF (Eq. \ref{grf_ts}, \ref{gz}, \ref{rgx}), except that LRF is built upon a local point set while GRF is based on the entire point cloud.
By computing the transformation pose, %$\{\rot_L^m,\,\trans_L^m,\,\size_L^m\}$, 
we calculate local point descriptors as 
\begin{equation}
\begin{aligned}
\label{eqn:lrf}
    % \mathcal{Q}_{L}^m &= \mathcal{\tilde{Q}}_{L}^m = \rot_L^\top (\mathbf{\tilde{q}}_{j} - \trans_L) / \size_L, 
    \mathcal{Q}_{L}^m &= \mathcal{\tilde{Q}}_{L}^m = \{({\rot_L^m})^\top (\mathbf{\tilde{q}}_{j} - \trans_L^m) / \size_L^m\}_{j=1}^{N_D}, 
    % \mathcal{\hat{Q}}_{BG}^m &=  (\mathbf{\hat{q}}_{j} - \trans_L) / \size_L, \\
    \\ 
    \qrypl &= \{\mathcal{Q}_{L}^m\}_{m=1}^{N^f}. 
    % \qrybg = \{ \mathcal{\hat{Q}}_{BG}^m\}_{m=1}^{N^f},
\end{aligned}
\end{equation}
Similarly, we transform $\refpf$ to $\refpl$.
Note that $\mathcal{Q}_{L}^m = \mathcal{\tilde{Q}}_{L}^m$ since LRF is pose-invariant. 
Then, the LRF encoding is extracted from $\qrypl$ and $\refpl$ with a three-layer MLP.
% we encode the LRF descriptors with three MLP layers.
 %The LRF descriptors focus on fine-level 
% Moreover, we kept the rotation-variant intermediate point clouds $\qrybg$ and $\refbg$ for two considerations.
% Firstly, after transforming by initial pose prediction $\diffp_{init}$, the rotation difference between $\qryphat$ and $\refp$ is relatively small. 
% Secondly, keeping original rotational information is important in fine-level matching.

% Though LRF provides detailed geometric information, it naturally lacks global information.
% To tackle this problem, we extract positional encoding for each local region from $\mathbf{\tilde{Q}}^{f}$ by leveraging a mini-PointNet~\cite{qi2017pointnet}.
% The positional encoding provides essential complementary global information to LRF encoding to establish accurate correspondences.

\begin{table*}[tbp]
    \centering
\tablestyle{5pt}{1.1}
\small
\begin{tabular}{@{}l | ccc | ccc | c |c@{}}
        % \shline 
        Method & Seg. Model & Modality & CAD model 
        & LM-O & TUD-L & YCB-V & Avg & Time (s) \\ 
        \shline 
        ZeroPose~\cite{chen2024zeropose} & SAM~\cite{kirillov2023segment} & RGB & \cmark & 35.6 & 42.1 & 53.4 &  43.7 & 3.90 \\
        CNOS~\cite{nguyen2023cnos}   & FastSAM~\cite{zhao2023fast} & RGB & \cmark & 39.7 & 48.0 & 59.9 &  49.2 & \textbf{0.23} \\
        CNOS~\cite{nguyen2023cnos}       & SAM~\cite{kirillov2023segment} & RGB & \cmark & 39.6 & 39.1 & 59.5 &  46.1 & 1.71 \\
        % SAM-6D~\cite{lin2023sam6d}   & FastSAM~\cite{zhao2023fast} & RGB & \cmark & 40.6 & 50.1 & 60.6 &  50.4 \\
        % SAM-6D~\cite{lin2023sam6d}   & SAM~\cite{kirillov2023segment} & RGB & \cmark & \underline{44.4} & 49.8 & 59.5 &  51.2 \\
        SAM-6D~\cite{lin2023sam6d}& FastSAM~\cite{zhao2023fast} & RGB-D  & \cmark & \underline{42.2} & 51.7 & 62.1 &  52.0 & 1.47 \\
        SAM-6D~\cite{lin2023sam6d}    & SAM~\cite{kirillov2023segment} & RGB-D  & \cmark & \textbf{46.0} & \textbf{56.9} & 60.5 &  \textbf{54.5} & 4.53 \\
        \hline 
        \textbf{UNOSeg (Ours)} & FastSAM~\cite{zhao2023fast}             & RGB  & \xmark & 37.3 & 50.3 & \textbf{67.3} &  51.6 & \underline{0.24} \\
        \textbf{UNOSeg (Ours)} & SAM~\cite{kirillov2023segment}                 & RGB  & \xmark & 39.7 & \underline{56.2} & \underline{66.8} &  \underline{54.2} & 2.68 \\
        % \shline 
\end{tabular}
\caption{  \label{tab:uno_seg}
    {\bf Segmentation results (mAP)
    %of existing methods
    on LM-O, TUD-L, and YCB-V.}
    Results of other methods are obtained from the BOP website {\href{https://bop.felk.cvut.cz/leaderboards/segmentation-unseen-bop23}{bop.felk.cvut.cz/leaderboards/segmentation-unseen-bop23}}.
    We denote the best score in \textbf{bold} and the second best score with \underline{underline}.
}  
\vspace{-1mm}
\end{table*}

\begin{table*}[tbp]
    \centering
\tablestyle{8pt}{1.1}
% \small
\begin{tabular}{@{}l | ccc | ccc | c | c@{}}
        % \shline
        Method & Seg. Model & Modality & Ref. Type
        & LM-O & TUD-L & YCB-V & Avg & Time (s) \\
        \shline
        ZTE-PPF~\cite{zteppf} & MRCNN~\cite{maskrcnn} & D & Model                      & \textbf{66.3} & 90.4 & 50.2 & 69.0 & \textbf{0.74} \\
        Koenig-PPF~\cite{konig2020hybrid} & Retina/MRCNN & RGB-D & Model  & 63.1 & \textbf{92.0} & \textbf{70.1} & \textbf{75.1} & 0.99 \\
        % SAM-6D~\cite{lin2023sam6d}  & SAM &  RGB-D &  Model & \textbf{69.9} & 90.4 & \textbf{84.5} & \textbf{81.6} & 1.9 \\
        \hline
        Ref. AlignCenter  & UNOSeg-SAM & D & Image               & 27.3 & 28.8 & 47.5 & 34.5 & 2.69 \\
        FPFH+RANSAC~\cite{rusu2009fast} & UNOSeg-SAM & D & Image & 31.0 & 31.0 & 50.0 & 37.3 & 6.38 \\
        FPFH+MAC~\cite{zhang20233d} & UNOSeg-SAM & D & Image     & 22.5 & 22.1 & 49.6 & 31.4 & 136.94 \\
        PPF~\cite{drost2010model} & UNOSeg-SAM & D & Image        & 29.7 & 14.8 & 38.3 & 27.6 & 11.79 \\
        PPF\_3D\_ICP~\cite{drost2010model} & UNOSeg-SAM & D & Image  & 44.7 & 29.1 & 66.8 & 46.9 & 14.27 \\
        FCGF+RANSAC~\cite{choy2019fcgf} & UNOSeg-SAM & D & Image & 38.9 & 59.0 & 57.6 & 51.8 & 10.96 \\
        FCGF+MAC~\cite{zhang20233d} & UNOSeg-SAM & D & Image     & 33.9 & 48.3 & 51.0 & 44.4 & 60.53 \\
        UTOPIC \cite{chen2022utopic} & UNOSeg-SAM & D & Image    & 13.7 & 35.4 & 10.5  & 19.9 & 4.00 \\
        GeDi~\cite{poiesi2022gedi}   & UNOSeg-SAM & D & Image      & 42.8 & 67.3 & 60.6 & 56.9 & 48.89 \\
        FreeZe~\cite{andera2023freeze}& UNOSeg-SAM & RGB-D & Image& 45.5 & 68.3 & 65.5 & 59.8 & 52.96 \\
        SAM-6D$^*$~\cite{lin2023sam6d}  & UNOSeg-SAM & RGB-D & Posed Image & 54.5 & 29.7  &  68.1 & 50.8 & 4.21  \\
        %\hline
        \textbf{UNOPose (Ours)} & UNOSeg-FastSAM &  RGB-D & Image  & 55.8 & 66.5 & 81.9 & 68.1 & \textbf{1.11} \\
        \textbf{UNOPose (Ours)} & UNOSeg-SAM    &  RGB-D & Image  & \textbf{58.7} & \textbf{71.0} & \textbf{83.1} & \textbf{70.9} &  3.70 \\
        % \shline
\end{tabular}
\caption{\label{tab:uno_pose_sota}
{\bf Pose estimation results 
% of 
% existing methods
on LM-O, TUD-L, and YCB-V.}
The mean Average Recall (\%) of the BOP metric and the average time (s) per image are reported.
We highlight the best scores of each setting in \textbf{bold}.
Specifically, for adapting SAM-6D~\cite{lin2023sam6d} to our setting, we randomly choose one of the two templates which \cite{lin2023sam6d} uses as a canonical reference during training.
}
\vspace{-1mm}
\end{table*}

Positional encoding and LRF encoding are combined with the geometric and color features as the input of the Geometric Transformer.
% We incorporate these local point descriptors into LRF encoding to provide fine-grained structural details for fine-level point matching.
% 
% The geometric and color features are extracted similarly to the coarse point matching procedure, but we extend a branch to include positional encoding for the color features.
% Concretely, it is extracted from LRF local descriptors and ball-grouping local region sets $\qrybg, \refbg$ (as in~\cite{qi2017pointnet++}) using a mini-PointNet~\cite{qi2017pointnet}.
% We incorporate positional encodings, extracted from local descriptors using a mini-PointNet~\cite{qi2017pointnet}, into color features.
Similar to initial pose prediction, by decoding these features,
we obtain the fine point-wise features $\hat{F}^f_Q, \hat{F}^f_P$ and overlap confidences $\hat{O}^f_Q, \hat{O}^f_P$, which are then used to obtain the overlap-aware fine correlation matrix $\mathbf{X}^{f} \in \mathbb{R}^{(N^{f} + 1)\times (N^{f} + 1)}$. 
The final pose $\diffp$ is predicted by solving \eqn{eq:target} using $\mathbf{X}^{f}$ with the weighted SVD algorithm.

\section{Experiments}
\label{sec:exp}
\begin{table*}[t]
    \centering    
    %%%%%%%%%%%%%%%%%%%%%%%%%%
    \begin{subfigure}[t]{0.49\linewidth}
    \vspace{6pt}
    \tablestyle{14pt}{1.5}
	\begin{tabular}{@{}l | ccc | c@{}}
	% \shline
        Category & VSD & MSSD & MSPD & $\text{AR}_{\text{BOP}}$ \\
        \shline
        Dragon & 47.1  & 73.2 &  69.4 &  63.2\\
        Frog   & 59.4  & 52.6 &  57.6 &  56.4\\
        Can    & 56.8  & 65.6 &  65.2 &  62.5\\
        \hline
        Avg    & \textbf{54.4}  & \textbf{63.7} &  \textbf{64.1} &  \textbf{60.7}\\
    % \shline
    \end{tabular}    
    \vspace{0pt}
    \caption{\label{tab:onerefpercat}}
    \end{subfigure}~~~~\hfill
    %%%%%%%%%%%%%%%%%%%%%%%%%%%%%%%%%%%
    \begin{subfigure}[t]{0.49\linewidth}
        \vspace{0pt}
        \includegraphics[width=0.96\linewidth]{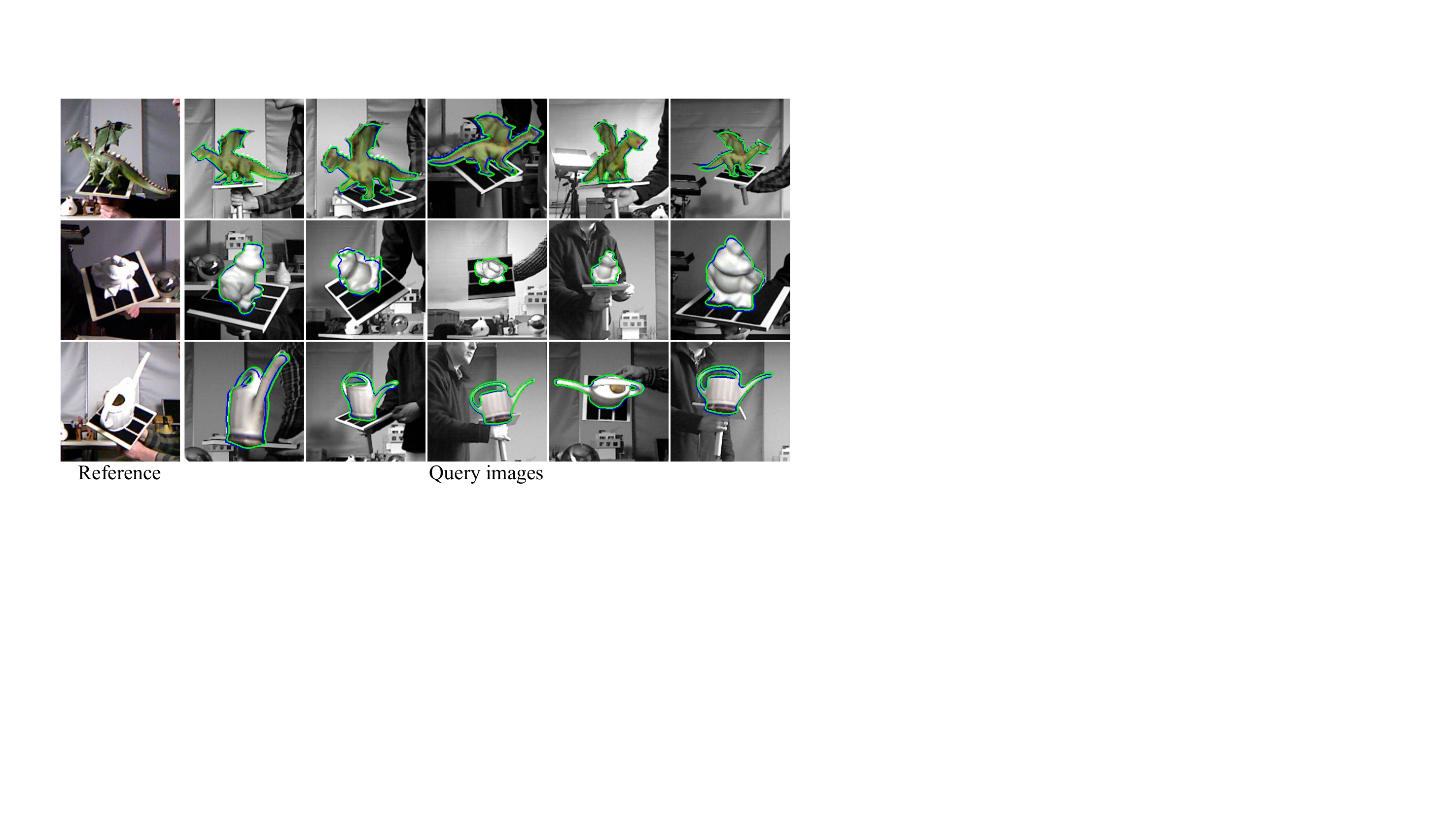}
        \caption{\label{fig:onerefpercat}}
    \end{subfigure}\hfill
    \caption{
        {\bf One reference per category experiments on TUD-L.}
         \textcolor{blue}{Blue} and \textcolor{green}{green} contours denote GT and estimated poses, respectively.
    }
    %%%%%%%%%%%%%%%%%%%%%%%%%%%%%%%%%%%
% \captionlistentry[figure]{figure and table}
% \captionsetup{labelformat=andtable}
\label{tab:oneref}
\vspace{-2mm}
\end{table*}

%%%%%%%%%%%%%%%%%%%%%%%%%%%%%%%%%%%%%%%%%%%%%%%%%%%%%%%%%%
\subsection{Experimental Setup}
\textbf{Implementation Details.}~~
Implemented with PyTorch~\cite{paszke2019pytorch},
we use the DINOv2~\cite{oquab2023dinov2,darcet2023vitneedreg} pre-trained ViT-base~\cite{dosovitskiy2020image} to encode color information. 
GeoTransformers~\cite{qin2023geotransformer} 
with attention \cite{attention} 
and sparse-to-dense linear attention \cite{lin2023sam6d} are respectively employed to extract coarse and fine geometric correspondences. 
The network is trained with standard MegaPose synthetic dataset~\cite{labbe2023megapose} for BOP unseen object pose estimation track~\cite{hodan2024bop}.
It contains $\sim$2M images of $\sim$50K objects from GSO~\cite{downs2022google} and ShapeNet~\cite{chang2015shapenet}.
We train 3 epochs with a batch size of 32 on 4 RTX 3090 GPUs.
The network is trained with Adam~\cite{KingmaB_adam_iclr15} and cosine annealing \cite{cosine_lr}, using a base learning rate of $10^{-4}$.
Moreover, we leverage InfoNCE~\cite{oord2018representation} loss for learning correlation matrix and weighted BCE loss for supervising overlap prediction.

\nbf{Datasets and Benchmarks.}~~
We evaluate our method on BOP test split~\cite{hodan2020bop} of three different real-world datasets: 
LM-O~\cite{brachmann2014learning},  TUD-L~\cite{hodan2018bop}, and YCB-V~\cite{posecnn}.
\emph{LM-O} consists of 200 images from one clutter scene, where eight objects with occlusion are provided for testing.
\emph{TUD-L} comprises three moving objects with mild occlusion and diverse lighting conditions.
\emph{YCB-V} is very challenging due to severe occlusions and sensor noise. It comprises 900 testing images from 12 scenes consisting of 21 objects.
For each query object in the test split, we randomly select a reference object beyond the test set (in the training or evaluation set).
Unless specified, we constrain the maximal rotation difference between reference and target to 50$^\circ$ to ensure the ground-truth correspondence set is not empty.
Moreover, the occlusion ratio of reference objects is required to be less than 30\%.

\nbf{Evaluation Metrics.}~~
For evaluating instance segmentation, we follow \cite{hodan2020bop} to use the mean Average Precision (mAP) metric, which is computed across different Intersection-over-Union (IoU) thresholds (\{0.5, 0.55, ..., 0.95\}). 
Only annotated instances with at least 10\% projected visible surfaces are considered for evaluation.
For object pose evaluation, we use the symmetry-aware BOP metric~\cite{hodan2020bop}.
It is calculated as the mean of the Average Recall of Visible Surface Discrepancy (VSD), Maximum Symmetry-Aware Surface Distance (MSSD), and Maximum Symmetry-Aware Projection Distance (MSPD) (See \cite{hodan2020bop} for details).
% Please refer to~\cite{hodan2020bop} for a detailed explanation of these metrics.

%%%%%%%%%%%%%%%%%%%%%%%%%%%%%%%%%%%%%%%%%%%%%%%%%%%%%%%%%%
\subsection{UNO Object Segmentation and Pose Estimation Results}
\nbf{UNO Object Segmentation Results.}~~
We compare our UNOSeg with some existing CAD-model-based unseen object segmentation methods~\cite{chen2024zeropose,nguyen2023cnos,lin2023sam6d} in \tbl{tab:uno_seg}.
Leveraging a single image as a reference, our UNOSeg achieves comparable results with state-of-the-art methods (Ours 54.2\% \vs SAM-6D 54.5\%) at a faster speed (Ours 2.68s \vs SAM-6D 4.53s).
Notably, UNOSeg-FastSAM achieves the best performance on  YCB-V (mAP 67.3\%).
Compared to FastSAM~\cite{zhao2023fast}, SAM~\cite{kirillov2023segment} achieves consistently better results on three datasets.
Therefore, we choose UNOSeg-SAM as the default segmentation for subsequent experiments.

%%%%%%%%%%%%%%%%%%%%%%%%%%%%%%%%%%%%%%%%%%%%%%%%%%%%%%%%%%

\nbf{UNO Object Pose Estimation Results.}~~
% \subsubsection{Comparision with Existing Methods}
Since we are the first to conduct relative 6DoF pose estimation for unseen objects and build a brand-new benchmark, we re-implement some traditional~\cite{rusu2009fast,zhang20233d,drost2010model} and learning-based methods~\cite{chen2022utopic,poiesi2022gedi,andera2023freeze,choy2019fcgf,lin2023sam6d} for comparison.
\textbf{Note that we leverage identical data pre-processing, segmentation, query and reference image pairs, and evaluation protocols for all methods for fair comparison.}
Moreover, we also compare \netname~with two CAD-model-based pose estimators leveraging $SE(3)$-invariant PPF feature~\cite{zteppf,konig2020hybrid}.
All results are illustrated in \tbl{tab:uno_pose_sota}.
It clearly shows that our method surpasses all image-reference-based methods by a large margin, and even achieves comparable results with CAD-model-based methods using class-specific detectors~\cite{maskrcnn,lin2017focal} (Ours 70.9\% \vs ZTE-PPF~\cite{zteppf} 69.0\% \vs Koenig-PPF~\cite{konig2020hybrid} 75.1\%).
``Ref. AlignCenter'' directly uses the reference rotation as prediction, meanwhile leveraging the shift between query and reference centers as relative translation, so it can be regarded as the baseline of all methods.
Among all traditional descriptors, PPF with Iterative Closest Point (ICP) refinement~\cite{drost2010model} achieves the best results (AR 46.9\%).
Learning-based methods demonstrate varying generalization abilities on unseen objects.
For example, the GeoTransformer-based method UTOPIC, designed for registering object-level partial point clouds, performs suboptimally in our setting.
In contrast, the adapted unseen object pose estimation methods, \ie SAM-6D~\cite{lin2023sam6d} and FreeZe~\cite{andera2023freeze}, achieve satisfactory results compared to their counterparts.
% Notably, FreeZe~\cite{andera2023freeze}, which employs Gedi~\cite{poiesi2022gedi} to extract geometric features and DINOv2~\cite{oquab2023dinov2} to extract color features, achieves the closest performance to our \netname~(Ours 70.9\% \vs FreeZe 59.8\%).

\nbf{Run-time Analysis.}~~
Evaluated on a single RTX 3090 GPU, our pipeline runs at 3.70s for one image, including 2.68s for segmentation and 1.02s for pose estimation (Tab.~\ref{tab:uno_seg} - \ref{tab:uno_pose_sota}).

\subsection{One Reference for a Category}
To minimize the efforts of acquiring references, we leverage a single reference for all objects of the same category in a dataset.
\tbl{tab:oneref} illustrates the qualitative and quantitative experimental results on TUD-L.
It is worth noting that the target object in TUD-L is dynamic rather than static, preventing methods from leveraging background information. This setup poses challenges for relative pose estimation methods that rely on scene features, making them likely to fail in this task. Despite this,
\netname~achieves impressive performance (60.7\% in terms of the $\arbop$ metric) with only a single reference per category.
The visualization results in \tbl{fig:onerefpercat} further show \netname~is capable of predicting omnidirectional relative poses.
Additional experiments on reference selection are provided in the appendix.

\begin{table}[t]    
    \centering
    %%%%%%%%%%%%%%%%%%%%%%%%%%
    \tablestyle{5.8pt}{1.1}
	\begin{tabular}{@{}l | ccc | ccc | c@{}}
	% \shline
        ROW & GRF & LRF & FPE & VSD & MSSD & MSPD & $\text{AR}_{\text{BOP}}$ \\
        \shline
        A0 & \xmark & \xmark & \xmark & 64.6 & 68.1 & 59.2 & 63.9 \\
        A1 & \cmark & \xmark & \xmark & 66.4 & 72.8 & 63.2 & 67.5 \\
        A2 & \xmark & \cmark & \xmark & 66.2 & 69.5 & 60.4 & 65.4 \\
        A3 & \cmark & \cmark & \xmark & 66.9 & 73.2 & 63.5 & 67.9 \\
        B0 & \xmark & \xmark & \cmark & 79.8 & 85.0 & 75.5 & 80.1 \\
        B1 & \cmark & \xmark & \cmark & 79.4 & 85.2 & 76.6 & 80.4 \\ 
        B2 & \xmark & \cmark & \cmark & \textbf{81.1} & 85.1 & 76.6 & 80.9 \\
        B3 & \cmark & \cmark & \cmark & 80.7 & \textbf{86.0} & \textbf{77.7} & \textbf{81.5}   \\
    % \shline
    \end{tabular}    
    \caption{\label{tab:pose_arch_abla}
        \textbf{Ablation of GRF and LRF on YCB-V.} FPE stands for ``fine pose estimation''.
    }
\end{table}

\subsection{Ablation Studies}
% We ablate some key ingredients in \netname~and present the results in \tbl{tab:pose_arch_abla}, \tbl{tab:pose_other_abla} and \fig{fig:ref_rot_err}.

% \textbf{Ablation on Network Architecture.}~~
% We follow the broadly used coarse-to-fine refinement paradigm to design our network.
% It is shown that the fine point matching procedure raises the performance distinctly (\tbl{tab:pose_abla} A1 \vs A0).
% Moreover, ViT pre-trained by DINOv2~\cite{oquab2023dinov2} shows better generalization ability than MAE~\cite{he2022masked} in our scenario (\tbl{tab:pose_abla} A2).
\begin{table}[t]    
    \centering
    %%%%%%%%%%%%%%%%%%%%%%%%%%
    \tablestyle{3pt}{1.1}
	\begin{tabular}{@{}l | l | ccc | c@{}}
	% \shline
        ROW & Method & VSD & MSSD & MSPD & $\text{AR}_{\text{BOP}}$ \\
        \shline
        C0 & B3 + Soft Correspondence~\cite{yu2021cofinet} & 77.1 & 81.7 & 72.7 & 77.2 \\
        C1 & B3 + Overlap Predictor & \textbf{82.6} & \textbf{87.4} & \textbf{79.4} & \textbf{83.1} \\
        C2 & A3 + Overlap Predictor & 68.4 & 74.6 & 64.8 & 69.2 \\
        \hline 
        D0 & C1: Seg. $\rightarrow$ CNOS~\cite{nguyen2023cnos} & 77.4 & 80.9 & 73.6 &  77.3 \\
        D1 & C1: Seg. $\rightarrow$ SAM-6D~\cite{lin2023sam6d} & 78.3 & 82.1 & 74.3 &  78.2 \\
        D2 & C1: Seg. $\rightarrow$ MRCNN~\cite{maskrcnn,labbe2020cosypose} & 81.2 & 85.7 & 77.9 & 81.6 \\
        D3 & C1: Seg. $\rightarrow$ GT Seg. & \textbf{86.6} & 89.2 & 82.1 & \textbf{86.0} \\
        \shline
        E0 & C1: Ref. Real $\rightarrow$ Rendered & 83.7 & \textbf{89.6} & \textbf{83.1} & 85.5 \\
    % \shline
    \end{tabular}    
    \caption{\label{tab:pose_other_abla}\textbf{Ablation of overlap predictor, segmentation, and reference type on YCB-V.}}
\end{table}

\begin{figure}[t]
\centering
\includegraphics[width=0.9\linewidth]{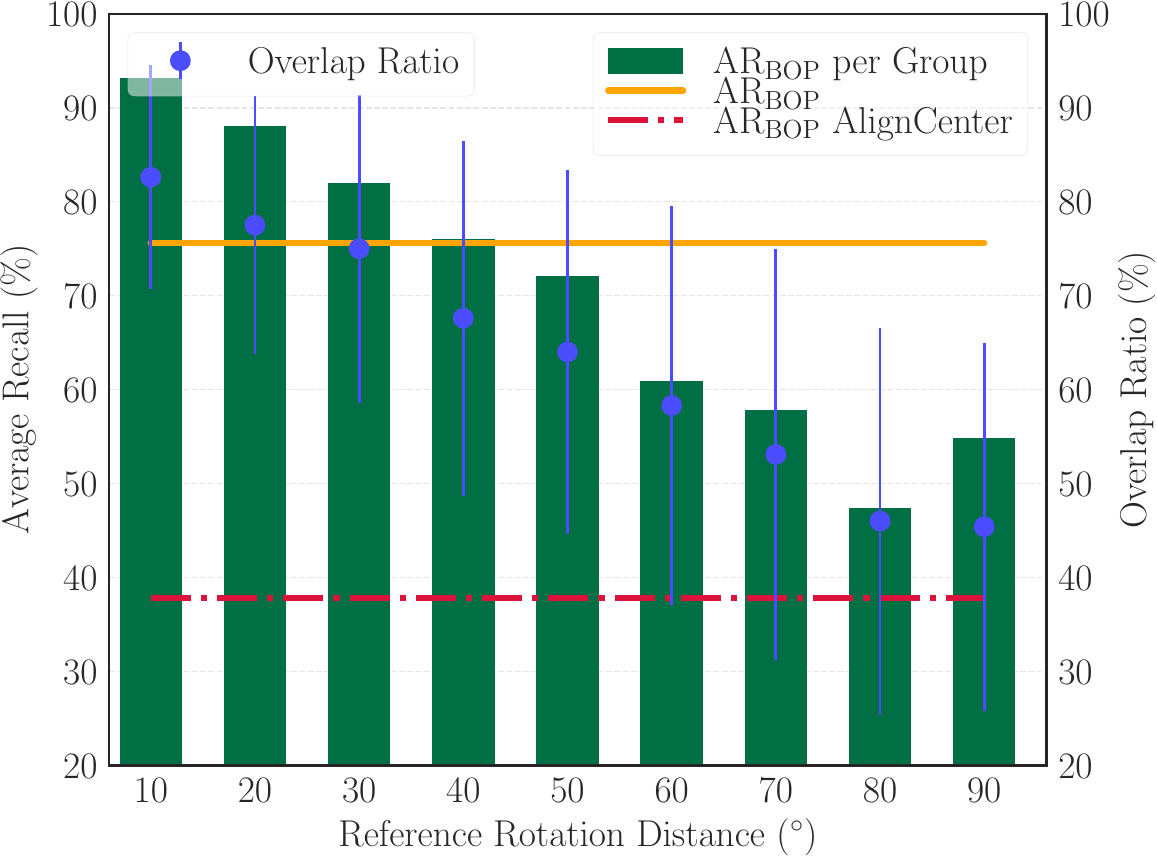}
\caption{\label{fig:pose_abla} \textbf{Ablation of initial rotation distance between query and reference objects.} We categorize all testing objects into nine groups according to the initial rotation distance, and evaluate the $\arbop$ metric and overlap ratio for each group separately.
}
\label{fig:ref_rot_err}
\end{figure}

\nbf{Effectiveness of GRF and LRF.}~~
As demonstrated in \sect{sec:method_pose},
GRF transfers arbitrary object poses to an \(SE(3)\)-invariant frame, while LRF assists with extracting precise local descriptors for fine pose estimation.
%GRF is for transferring an arbitrary object pose to an $SE(3)$-invariant frame, 
%while LRF is for extracting precise local descriptors that benefit fine point matching.
\tbl{tab:pose_arch_abla} shows that the effectiveness of GRF and LRF is complementary in UNOPose.
Specifically, GRF brings significant performance enhancement in the coarse pose estimation (\tbl{tab:pose_arch_abla} A1 \vs A0, A3 \vs A2), providing an accurate pose initialization for the next stage.
Meanwhile, LRF is constructed to exploit fine-grained local information, primarily enhancing the fine pose estimation results (\tbl{tab:pose_arch_abla} B0 \vs B2, B1 \vs B3).
With both GRF and LRF, our network achieves obvious performance improvement (\tbl{tab:pose_arch_abla} A3 \vs A0, B3 \vs B0).
% Both designs are important in our network (\tbl{tab:pose_arch_abla} B0, B1).
% Moreover, with the combination of GRF and LRF, the network achieves a more accurate prediction (\tbl{tab:pose_arch_abla} B2 \vs A2).

\nbf{Ablation on Correspondence Loss.}~~
% \tbl{tab:pose_abla} (C0 \vs B3) shows distinct improvement with the introduction of overlap prediction.
In partial-to-partial point cloud registration, only some of the points can find their counterparts on the target point cloud.
Therefore, the introduction of \textbf{overlap prediction} helps to distinguish the points in the overlap region and achieves good results (Tab.~\ref{tab:pose_arch_abla}-\ref{tab:pose_other_abla} C1 \vs B3, C2 \vs A3).
Moreover, we consider soft correspondence loss in~\cite{yu2021cofinet} as an alternative, which predicts probabilities of overlap between each point pair.
However, the result did not meet the expectations (\tbl{tab:pose_other_abla} C0).

\nbf{Ablation on Segmentation.}~~
\tbl{tab:uno_seg} showcases the advantage of our UNOSeg over other segmentation methods on YCB-V.
Furthermore, substituting UNOSeg with CNOS~\cite{nguyen2023cnos}, SAM-6D~\cite{lin2023sam6d}, or MRCNN~\cite{maskrcnn,labbe2020cosypose} results in a noticeable decline in UNOPose's performance
%drops in varying degree 
(\tbl{tab:pose_other_abla} D0--D2).
Moreover, substituting UNOSeg with ground-truth segmentation brings $\approx$3\% performance enhancement (\tbl{tab:pose_other_abla} D3 \vs C1).
% Furthermore, when replacing UNOSeg with CNOS~\cite{nguyen2023cnos} and SAM-6D~\cite{lin2023sam6d}, UNOPose's performance drops in varying degree (\tbl{tab:pose_abla} F0, F1).
% Notably, UNOSeg even outperforms the dataset-specific detector MRCNN~\cite{maskrcnn,labbe2020cosypose}.

\nbf{Using Rendered References.}~~
We replaced the real reference with a pose-identical rendered version and present the results in \tbl{tab:pose_other_abla}.
With reduced depth noise and domain gap between training and testing, UNOPose works even better using the rendered reference (\tbl{tab:pose_other_abla} E0 \vs C1).
However, this ablation requires the object model, which is more difficult to acquire than an unposed real image.

\nbf{Ablation on Reference Rotation Distance.}~~
To investigate the impact of the initial rotation distance between the query and reference images,
we randomly select the reference object with a maximum rotation distance of 90$^\circ$ for each query object.
Further, all testing objects are categorized into nine groups according to initial rotation distances.
We separately evaluate each group and present the results in \fig{fig:ref_rot_err}.
Meanwhile, we calculate the overlap ratio between reference and query instances of each group and visualize the mean and standard deviation.
Notably, performance significantly decreases when the rotation distance exceeds 50$^\circ$. 
This decline is attributed to the reduced overlap between query and reference viewpoints.
However, our network still shows favorable transfer ability in extreme relative pose estimation (rotation distance (80$^\circ$, 90 $^\circ$], overlap ratio 45.4\%, $\arbop$ 54.8\%). % thanks to the overlap predictor.

The appendix presents further implementation details, visualization results, and more experimental analysis on reference selection and extensive datasets.
\section{Conclusion, Limitation and Future work}
\label{sec:conclusion}
This work has introduced \netname, a new approach for unseen object pose estimation with one reference.
% The key idea is fully exploiting $SE(3)$-invariant geometric features 
The key idea is constructing the $SE(3)$-invariant 
%global and local 
reference frame
%and $SE(3)$-consistent visual features 
for tackling diverse pose and size variations.
Moreover, we propose an overlap predictor for handling low-overlap scenarios.
To evaluate the proposed method, we devise a new benchmark based on the BOP challenge and compare some state-of-the-art methods. % traditional and learning-based approaches.
Experimental results show \netname~surpasses all compared reference-based methods significantly,
%by a large margin
% and is competitive with CAD-model-based state-of-the-art methods.
and is competitive with some \(SE(3)\)-invariant-feature-based methods relying on CAD models.

\nbf{Limitation and Future Work.} While efficient, correspondence-based methods rely on predicting robust and dense correspondence between query and reference objects, thereby limited in extreme two-view geometry applications.
Future work will focus on reconstructing the unseen object from a single reference~\cite{liu2023zero123,zou2024triplane} and concurrently estimating its object pose. 

\smallskip
\nbf{Acknowledgements.} This work was supported in part by the National Natural Science Foundation of China under Grant No.~62406169, 
and in part by the China Postdoctoral Science Foundation under Grant No.~2024M761673. 
% max 8 pages for submission

{
    \small
    \bibliographystyle{ieeenat_fullname}
    \bibliography{main}
}


\end{document}

\label{sec:supp}
% Optionally include supplemental material (complete proofs, additional experiments and plots) in appendix.
% All such materials \textbf{SHOULD be included in the main submission.}

\renewcommand\thefigure{\thesection-\arabic{figure}}
\renewcommand\thetable{\thesection-\arabic{table}}
\renewcommand\theequation{\thesection-\arabic{equation}}
\setcounter{figure}{0} 
\setcounter{table}{0}
\setcounter{equation}{0}

\maketitle
\appendix

\tableofcontents

\section{Additional Details of Method}

\subsection{UNO Object Segmentation}
% UNOSeg is constructed with the segmentation stage and matching stage.
UNOSeg is constructed with three steps, \ie, mask proposal generation, global and local matching, and mask proposal assignment.

\paragraph{Mask Proposal Generation.}  
Given the query image $I_q$ and uniformly sampled pixel positions $P$ as prompt, the segment anything model (SAM) \cite{kirillov2023segment,zhao2023fast} $\Theta$ predicts $N_m$ mask proposals $\mathcal{M}$ with confidence scores $\mathcal{C}$, denoted by
\begin{equation}
\label{maskproposal}
    \mathcal{M}, \mathcal{C} = \Theta(I_q, P).
\end{equation}
We discard low-confidence predictions and apply Non-Maximum Suppression to filter duplicate proposals.
% We remove predictions that are either of low confidence or small in size by applying thresholds, and utilize Non-Maximum Suppression to filter repetitive instances.

\paragraph{Global and Local Matching.} 
In the matching stage, the network assigns each mask proposal a similarity score \wrt the reference view.
Specifically, given the reference image, we first remove the background using $M_p$, then crop the region of interest and resize it to $\hat{I}_p \in \mathbb{R}^{224 \times 224 \times 3}$.
Concurrently, we crop and resize the target image with all mask proposals to a consistent size $\{\hat{I}_q^j \in \mathbb{R}^{224 \times 224 \times 3}\}_{j=1}^{N_m}$.
$\hat{I}_q$ and $\hat{I}_p$ are fed into a pre-trained DINOv2 model \cite{oquab2023dinov2,darcet2023vitneedreg} to generate image-level global descriptors $\hat{\mathcal{G}}_q, \hat{\mathcal{G}}_p$ and $N_l$ patch-level local descriptors $\{\hat{\mathcal{L}}_q^k\}_{k=1}^{N_l}, \{\hat{\mathcal{L}}_p^k\}_{k=1}^{N_l}$. 
By evaluating the cosine similarity of descriptors, the matching score $\xi$ can be obtained as
\begin{equation}
\xi = (\xi_\mathcal{G} + \xi_\mathcal{L}) / 2, 
\end{equation}
where $\xi_\mathcal{G}$ and $\xi_\mathcal{L}$ are global and local descriptor similarities calculated by
\begin{equation}
\begin{aligned}
%\text{where} \quad 
\xi_\mathcal{G} &= \frac{\hat{\mathcal{G}}_q^\top \hat{\mathcal{G}}_p}{\|\hat{\mathcal{G}}_q\|_2 \cdot \|\hat{\mathcal{G}}_p\|_2 }, \\
 \quad 
\xi_\mathcal{L} &=  \frac{1}{N_l} \sum_{k=1}^{N_l} 
\max_{i=1,...,N_l}
\frac{ (\hat{\mathcal{L}}_q^k)^\top \hat{\mathcal{L}}_p^i}{\|\hat{\mathcal{L}}_q^k\|_2 \cdot \|\hat{\mathcal{L}}_p^i\|_2}.
\end{aligned}
\end{equation}
% Here $<,>$ denotes the inner product.
Leveraging both global and local matching scores, the network effectively distinguishes the mask proposal most similar to $M_q$ relative to the reference object.

\paragraph{Mask Proposal Assignment.} 
A single query image may contain multiple distinct query objects in the test scenario.
In this scenario, we generate mask proposals using \eqn{maskproposal} once.
For each individual mask proposal, we calculate its similarity score against every candidate reference and select the highest score to determine the object class of this mask proposal.
Noteworthy, while the whole image may have multiple reference images, each distinct target object only has a single reference image.

\subsection{Details of Eq. (5) in GRF Construction}
Here we provide a detailed definition of $\rrot_{Gx}$ in the construction of GRF (\cf Eq. (5) in the main text).
Considering the vector from the center $\mathbf{c}_Q$ to a point $\mathbf{q}\in\mathbf{Q}_{cam}$, given the normal vector $\rrot_{Gz}$ at $\mathbf{c}_Q$, its projected vector $\mathbf{v_q}$ on the tangent plane of $\mathbf{c}_Q$ can be computed as
\begin{equation}
    \mathbf{v_q} = (\mathbf{q} - \mathbf{c}_Q) - \rrot_{Gz}^\top (\mathbf{q} - \mathbf{c}_Q) \rrot_{Gz}.
\end{equation}
Then, the projected vectors of all points are aggregated by summing them, with each vector being weighted according to the distance between $\mathbf{q}$ and $\mathbf{c}_Q$.
We decrease the weight of points that are farther from the center, as they are more likely to be outliers.
Specifically, $\rrot_{Gx}$ is calculated as
\begin{equation}
\begin{aligned}
    % \rrot_{Gx} &= \sum_{\mathbf{q} \in \qryp} \frac{w_{q}^1 w_{q}^2 \mathbf{v_q}}{\|\sum_{\mathbf{q} \in \qryp} w_{q}^1 w_{q}^2 \mathbf{v_q}\|}, \\
    % \rrot_{Gx} &= \frac{1}{\|\sum_{\mathbf{q} \in \qryp} w_{q,1} w_{q,2} \mathbf{v_q}\|_2} \sum_{\mathbf{q} \in \qryp} w_{q,1} w_{q,2} \mathbf{v_q}, \\
    \rrot_{Gx} &= \frac{\sum_{\mathbf{q} \in \qryp} w_{q,1} w_{q,2} \mathbf{v_q}}{\|\sum_{\mathbf{q} \in \qryp} w_{q,1} w_{q,2} \mathbf{v_q}\|_2}, \\
    w_{q,1} &= (s_Q - \| \mathbf{q} - \mathbf{c}_Q \|_2)^2, \\
    w_{q,2} &= (\rrot_{Gz}^\top (\mathbf{q} - \mathbf{c}_Q))^2, \\
    s_Q &= \max_{\mathbf{q} \in \qryp} \| \mathbf{q} - \mathbf{c}_Q \|_2.
\end{aligned}
\end{equation}
The weighting factor $w_q$ in Eq. (5) can thus be expressed as
\begin{equation}
    w_q = \frac{w_{q,1} w_{q,2}}{\|\sum_{\mathbf{q} \in \qryp} w_{q,1} w_{q,2} \mathbf{v_q}\|_2}.
\end{equation}

\subsection{Details of Training Objectives}
We leverage the InfoNCE~\cite{oord2018representation} loss to constrain the learning of the correlation matrix and the weighted binary cross-entropy (WBCE) loss for supervising overlap prediction.
In specific, given the correlation matrix $\mathbf{X}^{c} \in \mathbb{R}^{(N^{c} + 1)\times (N^{c} + 1)}$ denoting the predicted correspondence between the query point cloud $\qrypgc$ and the reference point cloud $\refpgc$, it is supervised with
\begin{equation}
    \mathcal{L}_{X}^c = \text{CE} (\mathbf{X}^{c}[1:,\, :], \bar{y}_q) + \text{CE} (\mathbf{X}^{c}[:,\, 1:]^\top, \bar{y}_p).
\end{equation}
Here $\bar{y}_q \in \mathbb{R}^{N^c}$ and $\bar{y}_p \in \mathbb{R}^{N^c}$ are the ground-truth correspondence for $\qrypgc$ and $\refpgc$.
% TODO: def of \hat{y}_q

The overlap predictions $\hat{O}_{Q}^{c} \in \mathbb{R}^{(N^{c} + 1) \times 1}$, and $\hat{O}_{P}^{c} \in \mathbb{R}^{(N^{c} +1) \times 1}$ are supervised with
\begin{equation}
    \mathcal{L}_{O}^c = \text{WBCE} (\hat{O}_{Q}^{c}, \bar{O}_{Q}^{c}) + \text{WBCE} (\hat{O}_{P}^{c}, \bar{O}_{P}^{c}),
\end{equation}
where $\bar{O}_{Q}^{c}$, and $\bar{O}_{P}^{c}$ are overlap labels for the query and reference point clouds, respectively. 
For each point, its ground-truth overlap $\bar{o}^c_i, i\in\{1,...,N^c+1\}$ is calculated by
\begin{equation}
    \bar{o}_i^{c} = \begin{cases}
        0, ~~~~~~ \text{if}~~i=1~~\text{or}~~d_{i, \text{min}} > \delta, \\
        1, ~~~~~~ \text{otherwise}.
    \end{cases}
\end{equation}
For the background token ($i=1$), the ground-truth overlap score is constant at 0.
Moreover, $d_{i, \text{min}}$ is the distance of the $i$-th point to the closest point in the counterpart point cloud under ground-truth transformation, and $\delta$ is a hyper-parameter which we set to 0.15.

Similarly, we calculated $\mathcal{L}_{X}^f$ and $\mathcal{L}_{O}^f$ for supervising the fine point matching procedure.

To sum up, the overall learning objective can be written as
\begin{equation}
    \mathcal{L} = \sum_{t \in \{1,2,3\}} (\mathcal{L}_{X}^{c,t} + \mathcal{L}_{O}^{c,t}) + \sum_{t \in \{1,2,3\}} (\mathcal{L}_{X}^{f,t} + \mathcal{L}_{O}^{f,t}).
\end{equation}
Here $t$ denotes the block index of the geometric transformer decoder.

% \subsubsection{Results of Reference from Different Scenes}  % YCB-V

% \input{supp_materials/uno_ref_abla}
% \subsubsection{Analysis on the Number of References} % TUD-L
% Following \tbl{tab:oneref} in the main text, we dig into the influence of references' numbers.
% As shown in \tbl{tab:ref_num}, the performance gets saturated when the total reference number of each object increases to 16.
% It indicates that our \netname~does not rely on multiple references at the dataset level.

\subsection{Details of Hyper-parameters}
We empirically set the number of coarse point samples $N^c$ to 196, and the number of fine point samples $N^f$ to 2048.
We sample $N_H=300$ pose hypotheses in coarse pose prediction.
In building the local reference frame, the number of neighborhoods $N_D$ is set to 64.
The dimensions for geometric embedding, color embedding, LRF encoding, and positional encoding are set to 256.

\section{More Experimental Results}

\begin{table}[t]    
    \centering
    %%%%%%%%%%%%%%%%%%%%%%%%%%
    \tablestyle{2pt}{1.35}
	\begin{tabular}{@{}l |c c| ccc@{}}
	% \shline
        Method & Modality & \tabincell{c}{In-dataset\\fine-tuning} & \tabincell{c}{Angular\\error (°)$\downarrow$} & $\text{Acc}_\text{30}$ (\%) $\uparrow$	& $\text{Acc}_\text{15}$ (\%)$\uparrow$	 \\
        \shline
        RelPose~\cite{zhang2022relpose} & RGB & \cmark& 58.3 & 26.1 & 7.0 \\
        RelPose++~\cite{lin2023relpose++} & RGB &\cmark& 	46.6 & 42.5 & 	15.8 \\
        3DAHV~\cite{zhao20233d} & RGB &\cmark& 41.7 & 61.5 & 29.9  \\
        DVMNet~\cite{zhao2024dvmnet}  & RGB&\cmark & 36.8 & - & -\\
        UNOPose* & RGB &\xmark& 49.1 & 50.0 & 19.1 \\
        UNOPose & RGB-D &\xmark& \textbf{23.9} & \textbf{84.2} & \textbf{81.1} \\
    % \shline
    \end{tabular}    
    \caption{\label{tab:lm}
        \textbf{Comparison with RGB-based relative pose estimation methods on LM.} The depth of UNOPose* is predicted by ZoeDepth~\cite{bhat2023zoedepth}.
    }
\end{table}

\input{supp_materials/wildrgbd_vis}

\subsection{UNO Pose Estimation Results on LM}
Although the setting of UNOPose is different from that of RGB-based relative pose estimation methods~\cite{zhao20233d,zhao2024dvmnet,zhang2022relpose,lin2023relpose++}, we include a comparison of their test split of the LM dataset \cite{hinterstoisser2012} in \tbl{tab:lm}. 
We report the rotation error and the accuracy with thresholds of 30$^\circ$ and 15$^\circ$. 
For comparing our UNOPose with them under RGB modality, we use ZoeDepth~\cite{bhat2023zoedepth} to predict metric depth from monocular data.
It is shown that our method achieves comparable results with relative 3DoF pose estimators even under RGB modality.
Moreover, when using ground-truth depth, our approach largely surpasses in terms of rotation metrics. 
While leveraging more advanced monocular metric depth estimation techniques could potentially yield better results, it is beyond the scope of this paper, and we leave it for future work.
Additionally, we can estimate translation besides rotation. 
Note that, for evaluating on LM, they need to first train on the synthetic data and then perform in-dataset fine-tuning on the real LM data by excluding the test objects. 
However, UNOPose is only trained on the synthetic dataset and then tested on several real-world datasets. 

\subsection{Results on Real-world Scenarios}
To verify the practicality of UNOPose, we adopt it in real-world phone-captured scenarios and present qualitative results in \fig{fig:wildrgbd_vis}.
Specifically, we run UNOPose on three sequences of daily objects (banana, detergent, and kettle) provided by the WildRGB-D dataset~\cite{xia2024rgbd}. 
This dataset is photographed in several in-the-wild scenarios by the front camera of an iPhone. 
For each sequence, one frame is chosen as the reference, and the rest frames are treated as queries. The SAM-based masks provided by the dataset are directly employed. 
For visualization, we use the 3D bounding box derived from the reconstructed object and the provided absolute camera poses. 
Note that our UNOPose requires no information from the 3D model or the camera poses. 
Qualitative results show that UNOPose can adjust well to in-the-wild scenarios, daily objects, varying lighting conditions, low-quality depth, and occasional occlusions.

\subsection{More Ablation studies}

\begin{table}[t]    
    \centering
    %%%%%%%%%%%%%%%%%%%%%%%%%%
    \tablestyle{6.5pt}{1.45}
	\begin{tabular}{@{}l |ccc |c@{}}
	% \shline
        Method & VSD & MSSD & MSPD	& $\arbop$	 \\
        \shline
        Ref. AlignCenter (baseline) & 31.7 & 43.4 & 19.5 & 31.5 \\
        UNOPose & 69.2 & 74.8 & 63.9 & 69.3 \\
    % \shline
    \end{tabular}    
    \caption{\label{tab:diffscene}
        \textbf{Reference from different scenes.}
    }
\end{table}

\paragraph{Reference from Different Scenes.}
Compared to previous methods, UNOPose cuts objects out of the background, eliminating reliance on scene context.
Our approach allows reference images to originate from diverse scenes and be captured by different cameras.
We additionally present the evaluation results for all references from different scenes on YCB-V \cite{posecnn} and show the results in \tbl{tab:diffscene}. 
Compared to the baseline, UNOPose achieves a notable 37.8\% improvement in accuracy, demonstrating its robust adaptability across changing scenes.

\input{supp_materials/tudl_8ref}
\begin{table}[t]    
    \centering
    %%%%%%%%%%%%%%%%%%%%%%%%%%
    \tablestyle{15pt}{1.35}
	\begin{tabular}{@{}l |ccc |c@{}}
	% \shline
        Image ID & VSD & MSSD & MSPD	& $\arbop$	 \\
        \shline
        751   & 50.4 & 74.0 & 69.5 & 64.6 \\
        3895  & 51.0 & 74.7 & 70.5 & 65.4 \\
        4329  & 56.1 & 75.2 & 72.4 & 67.9 \\
        5274  & 56.3 & 76.1 & 73.5 & 68.6 \\
        7268  & 54.7 & 76.7 & 73.2 & 68.2 \\
        10741 & 47.1 & 73.2 & 69.4 & 63.2 \\
        11430 & 45.8 & 67.3 & 63.4 & 58.8 \\
        9734  & 36.3 & 57.3 & 52.1 & 48.5 \\
    % \shline
    \end{tabular}    
    \caption{\label{tab:refimg}
        \textbf{Ablation of the reference selection on ``One Reference per Category'' experiments on the ``Dragon'' object of TUD-L.} 10741 is the image ID we chose in the paper.
    }
\end{table}

\paragraph{Selection of Single Reference.}
In the main text, we conduct the ``One Reference for a Category'' experiment. 
To explore the effect of reference selection on the results, we further conduct ablations on the dragon object from the TUD-L dataset \cite{hodan2018bop}. 
Specifically, we sampled 8 different viewpoints from TUD-L, ensuring a rotation difference of at least 40 degrees between any two viewpoints.
This sampling strategy allowed the references to cover the object's full range of perspectives.
The further ``one reference per category'' experimental results are shown in \fig{fig:tudl_8ref} and \tbl{tab:refimg}.
It is shown that image IDs 751, 3895, 4329, 5274, and 7268 achieve better results than the viewpoint we chose in the paper (image ID 10741, $\text{AR}_\text{BOP}$ 63.2). 
However, when using 11430 or 9734 as the only reference image of the category, they gain sub-optimal results. % This decline is attributed to dim lighting (ID 11430) or reduced visible region (ID 9734).
Qualitative and quantitative results show that the reference with brighter illumination and larger visible regions can lead to better results.

\begin{table}[t]
    \tablestyle{12pt}{1.4}
	\begin{tabular}{@{}l | ccc | c@{}}
	% \shline
        Num. Ref. & VSD & MSSD & MSPD & $\text{AR}_{\text{BOP}}$ \\
        \shline
        1        & 54.4  & 63.7 &  64.1  &  60.7   \\
        8        & 53.4  & 65.9 &  65.6  &  61.6   \\
        12       & 61.3  & 74.2 &  74.0  &  69.8   \\
        16       & 62.5  & 74.3 &  75.4  &  70.7   \\
        32       & \textbf{62.8}  & \textbf{76.2} &  \textbf{75.7}  &  \textbf{71.6}  \\
        42       & 60.7  & 75.0 &  74.8  &  70.0   \\
        Random   & 61.9  & 75.2 &  75.9  &  71.0   \\
    \end{tabular}
    \centering
    \caption{
        {\bf Ablation on the number of references \wrt TUD-L.}
        We randomly select a certain number of references for each category on the TUD-L training set (denoted as Num. Ref.).
        ``Random'' is the setting in main experiments, indicating randomly selecting a reference for each query object.
        The reference selection policy is the same as the main text.
    }
\label{tab:numref}
\end{table}

\paragraph{Impact of Reference Number.}
We further dig into the influence of references' numbers on the whole dataset and present the results in \tbl{tab:numref}. 
The number of references (Num. Ref.) that equals 1 corresponds to the experiment in Tab. 3 of the main text, and ``random'' indicates randomly selecting a reference for each query object. 
Note that \emph{UNOPose still runs with only one reference for each query}. 
We find that the best performance is achieved with 32 references and competitive results can be observed after 
% gets saturated when 
the total reference number of each object increases to 12. 
Impressively, even with just one reference for each object in the entire dataset, the results remain excellent (comparable to using 8 references).
It indicates that UNOPose does not rely on multiple references at the dataset level.

\input{supp_materials/supp_vis}
\subsection{Qualitative Results}
We show some qualitative results on YCB-V, LM-O, and TUD-L datasets in \fig{fig:supp_vis}. 
Despite occlusion, sensor noise, and varying testing scenarios, our generic approach achieves robust unseen object pose estimation results with one single reference image.

\section{Code}
We provide the code of \netname~and results of UNOSeg in folder ``UNOPose\_code\_submit'' in our supplementary material.
Detailed instructions can be found in README.md.
The code and dataset will be publicly available upon acceptance.

\section{Potential Positive and Negative Societal Impacts}
\label{sec:broader_impacts}
% https://cvpr.thecvf.com/Conferences/2025/EthicsGuidelines
We developed \netname~for estimating unseen object poses given a single RGB-D reference image. 
By avoiding the need for retraining and reducing the cost of creating references for each new object, \netname~can not only reduce the environmental burden but also offer more application possibilities. 
This technology has broad applications in industrial manufacturing and robotic manipulation. 
While it may lead to job displacement due to increased automation, we aim for this work to have a positive societal impact by enhancing efficiency and safety.

{
    \small
    \bibliographystyle{ieeenat_fullname}
    \bibliography{main}
}
